# *Roget's Thesaurus*: a Lexical Resource to Treasure

**Mario JARMASZ**, **Stan SZPAKOWICZ**
School of Information Technology and Engineering
University of Ottawa
Ottawa, Canada, K1N 6N5
{mjarmasz,szpak}@site.uottawa.ca

**Abstract**

This paper presents the steps involved in creating an electronic lexical knowledge base from the 1987 Penguin edition of *Roget's Thesaurus*. Semantic relations are labelled with the help of *WordNet*. The two resources are compared in a qualitative and quantitative manner. Differences in the organization of the lexical material are discussed, as well as the possibility of merging both resources.

## Introduction

Peter Mark Roget published his first *Thesaurus* (Roget, 1852) over 150 years ago. Countless writers, orators and students of the English language have used it. Masterman (1957) was the first to use it in natural language processing (NLP) to improve word-for-word machine translation. Others, such as Sparck Jones (1964), Morris and Hirst (1991), Yarowsky (1992), McHale (1998) and Ellman (2000) have used it for text classification, building lexical chains, word sense disambiguation and comparing the similarity of documents. Their use of the *Thesaurus* has been limited by the fact that an adequate machine tractable version is not available (Jarmasz and Szpakowicz, 2001a). Electronic versions do exist. The 1911 edition is available via *Project Gutenberg* (Hart, 1991) in text format. This edition is the source for the semantic network *Factotum* (Cassidy, 1996) as well as the Web thesauri *Thesaurus.com* (Lexico, 2001) and the one prepared by the *University of Chicago Project for American and French Treasury of the French Language* (Olsen, 1996). It has been shown that the 1911 edition is inadequate for NLP due to its limited and antiquated vocabulary (Stairmand, 1994).

We propose to formalize *Roget's Thesaurus* so as to have it in machine-tractable form (Wilks *et al.*, 1996:9). Sparck Jones (1964), Sedelow and Sedelow (1992) as well as Cassidy (1996) have all attempted to achieve this. As far as we know, we are the first to implement an electronic lexical knowledge base (ELKB) using an entire current edition.

Simply deciding what version of the *Thesaurus* to use as the source for the ELKB is not simple. There exist hundreds of reference manuals with the name *Roget's Thesaurus* in the title, yet only various editions of *Roget's International Thesaurus* (Chapman, 1977), Penguin's *Roget's Thesaurus of English words and phrases* (Kirkpatrick, 1998) as well as the 1911 edition have actually been used for NLP. We want to use a thesaurus that has a structure very similar to the original and an extensive up-to-date vocabulary. The 1911 and Penguin editions use a taxonomy like the one devised by Roget. The categories of the 1911 version can be viewed at *Thesaurus.com.* It contains 1,000 concepts. Although the vocabulary in *Roget's International Thesaurus* is very rich, Chapman decided to use an entirely different taxonomy than the original. We therefore decided to license for research the 1987 edition of Penguin's *Roget's Thesaurus* in text format, since it was the most adequate for our needs.

*WordNet* (Miller *et al.*, 1990) cannot be ignored when preparing an ELKB. It is one of the most widely used lexical resources in NLP. Language generation, word sense disambiguation, information retrieval and machine translation employ it. *WordNet*, however, is not flawless. For example, hardly any links exist between different parts of speech, proper nouns are not represented, there are few idiomatic expressions and the taxonomy has not been thoroughly planned. Such flaws could be avoided by integrating some lexical knowledge from *WordNet* into *Roget's Thesaurus*.

In this paper we reflect on the qualities of the ideal ELKB for NLP. We say which of these *Roget'* sand *WordNet* have, as well as what can be implemented by our software. The taxonomies, sizes and lexical material of both resources are compared. We will show that although conceptually similar, they are very different in design. We will discuss edge counting and labelling *Roget's* semantic relations with the help of *WordNet* as ways of representing the knowledge contained in the *Thesaurus*.

## 1 From *Roget's Thesaurus* to an Electronic Lexical Knowledge Base

Before attempting to create an ELKB it is important to identify its most desirable properties. We have decided to create a general-purpose lexical resource instead of a domain specific one. *WordNet* has proven that this type of resource can be useful. It remains to be shown whether *Roget's* can be similarly useful in NLP. Any good ELKB should have a large, modern vocabulary, a simple way of identifying the different word senses, a clear classification system, usage frequencies, several ways of grouping words and phrases to represent concepts, explicit links between the various units of meaning, for example between concepts, paragraphs or synsets, and an entire index of every single word and phrase in the resource. Additionally there should exist links between parts of speech, a lexicon containing idiomatic expressions and proper nouns as well as a simple way of extending the ELKB. Definitions and subcategorization information would also be beneficial.

Even though *Roget's Thesaurus* is a rich lexical resource, it is far from perfect. To the uninitiated or the unskilled, the *Thesaurus* can do more harm than good. It contains a rich vocabulary but does not explain how these words should be used, as a dictionary does, or what is the relation between these words other than that they are close in meaning. A common misconception is that *Roget's* is a dictionary of synonyms, yet we are warned that "… it is hardly possible to find two words having in all respect the same meaning, and being therefore interchangeable; that is admitting of being employed indiscriminately, the one the other, in all their applications". (Kirkpatrick 1998). The meanings of words and phrases in the *Thesaurus* are presumed to be known by the reader, but a computer system cannot be aware of these meanings. The only information that it can use to define a word is its location in the classification system and among the closely related words and phrases. This could be improved by mapping the various senses with definitions of an on-line dictionary but it is not a straightforward task. Words and phrases are placed into synset-like groups, called semicolon groups, that are contained in paragraphs found under heads that represent a concept. Kirkpatrick (ibid.) states "… words are grouped in semicolon groups according to their meaning, context or level of usage. Semicolon groups follow each other in a logical sequence, exploring every aspect of the idea under consideration." Although the *Longman's* corpus has been used for the preparation of the *Thesaurus*, it remains unclear if any frequency information is included in the ordering of words. Between them no explicitly labelled semantic relations exist. Finally, the index of the *Thesaurus* contains only about half of all its words and phrases.

The initial step in creating an ELKB from existing lexical material is cleaning up, expanding and reformatting the source. We have done this, as well as created an entire computer index and written a Java program to store and access the lexical material. We are currently implementing similarity measures based on edge-counting and labelling implicit semantic relations with the help of *WordNet*. Both of these tasks are discussed in Section 3. When possible, links will be added between parts of speech. *Roget's* potential lies in its classification system, described in the next section, and its rich vocabulary that contains some idiomatic expressions and proper nouns. In this phase of the project, we do not plan to calculate any frequency information, nor do we intend to map *Roget's* word senses onto dictionary definitions, although this mapping has been done on a small scale by Kwong (1998).

## 2 The difference between *Roget's Thesaurus* and *WordNet*

Thesauri are often described as "inverted dictionaries, which is to say, words organized so as to be found via their meanings, rather than, as in a standard dictionary, a list of meanings indexed by a word name" (Wilks *et al.*, 1996: 63). Miller wrote that "[t]he most ambitious feature of *WordNet*, however, is its attempt to organize lexical information in terms of word meanings, rather than word forms. In that respect, *WordNet* resembles a thesaurus more than a dictionary, and, in fact, Laurence

Urdang's revision of Rodale's *The Synonym Finder* (1978) and Robert L. Chapman's revision of *Roget's International Thesaurus* (1977) have been helpful tools in putting *WordNet* together". (Miller *at al.*, 1990) *Roget's Thesaurus* and *WordNet* seem to be very similar. If both were machine tractable, could they be used interchangeably for various NLP applications? We will try to answer this by comparing their ontologies, word counts and sets of semantic relations.

A major strength of *Roget's Thesaurus* is its unique system of classification "of the ideas which are expressible by language" (Roget, 1852). Our ELKB uses a slightly modified classification system that is organized in eight classes: *1. Abstract Relations*, *2. Space*, *3. Matter*, *4. Intellect: Formation of ideas*, *5. Intellect: Communication of ideas*, *6. Volition: Individual volition*, *7. Volition: Social volition*, *8. Emotion, religion and morality*. Roget devised this system in the following way: "I have accordingly adopted such principles of arrangement as appeared to me to be the simplest and most natural, and which would not require, either for their comprehension or application, any disciplined acumen, or depth of metaphysical or antiquarian lore" (ibid.). Within these eight classes are sections, and under the sections are 990 heads that represent various concepts. Within these heads, words and phrases are organized in semicolon groups, which belong to one of noun, adjective, verb, adverb or interjection paragraphs. This system of classification has withstood the test of time remarkably well, hardly changing in 150 years, a tribute to the robustness of *Roget's* design.

In *WordNet*, only nouns are clearly organized into a hierarchy. Adjectives, verbs and adverbs are organized individually into various webs that are difficult to untangle. This decision has been based on pragmatic reasons more than on theories of lexical semantics, as Miller (1998) admits: "Partitioning the nouns has one important practical advantage: it reduces the size of the files that lexicographers must work with and makes it possible to assign the writing and editing of the different files to different people." The noun hierarchies are organized around the following nine *unique beginners*: *1.{entity, something}, 2.{psychological feature}, 3.{abstraction}, 4.{state}, 5.{event}, 6.{act, human action, human activity}, 7.{group, grouping}, 8.{possession},9. {phenomenon}.*

*Roget's* ontology can be divided into classes that describe the external world (*Abstract Relations, Space, Matter)* and ones that describe the internal world of the human. The lexical material is almost evenly divided along these lines, 446 headwords belong to the external world, 544 to the internal world. *WordNet* seems to favour the external world, with only the unique beginners *{psychological feature}* and *{act, human action, human activity}* describing the internal world of the human. Intuition suggests that *Roget's* and *WordNet* should have a big overlap in lexical material pertaining to the material world. Table 1 presents the distribution of words and phrases within *Roget' s Thesaurus* ordered by class number. A list of about 45,000 strings that can be found in *WordNet 1.6* and the *1987 Roget's* has been identified. *% of c.h.*, *% of c.k.* and *% of c.s.* in Table 1 indicate the percentage of heads, keywords and strings respectively that can be found in this common word and phrase list. A keywords is the first word of a paragraph and serves to generally

| **Class #** | **# of sections** | **# of heads** | **# of paragraphs** | **# of SGs** | **# of strings** | **% of c.h.** | **% of c.k.** | **% of c.s.** |
|---|---|---|---|---|---|---|---|---|
| 1 | 8 | 182 | 1146 | 10479 | 38052 | 0.63 | 0.62 | 0.68 |
| 2 | 4 | 136 | 992 | 8904 | 32643 | 0.71 | 0.63 | 0.68 |
| 3 | 3 | 128 | 714 | 6447 | 22828 | 0.72 | 0.60 | 0.67 |
| 4 | 7 | 67 | 460 | 4243 | 15550 | 0.79 | 0.62 | 0.58 |
| 5 | 3 | 81 | 495 | 4953 | 18407 | 0.86 | 0.62 | 0.62 |
| 6 | 5 | 138 | 927 | 9626 | 36589 | 0.88 | 0.59 | 0.63 |
| 7 | 4 | 84 | 447 | 4125 | 15875 | 0.88 | 0.65 | 0.59 |
| 8 | 5 | 174 | 1251 | 11150 | 44870 | 0.84 | 0.56 | 0.57 |
| **Total:** | 39 | 990 | 6432 | 59927 | 224814 | 0.78 | 0.61 | 0.63 |

**Table 1:** Distribution of words and phrases within *Roget's Thesaurus* ordered by class number

| % of c.s. | Class # | Head | H. in WN | # of paragraphs | # of SGs | # of strings | % of c.k. |
|---|---|---|---|---|---|---|---|
| 0.97 | 1 | 42: Decrement: thing deducted | No | 1 | 10 | 36 | 1.00 |
| 0.97 | 1 | 77: Class | Yes | 5 | 28 | 121 | 0.80 |
| 0.94 | 5 | 567: Perspicuity | Yes | 2 | 9 | 31 | 1.00 |
| 0.87 | 3 | 402: Bang: sudden and violent noise | No | 4 | 25 | 86 | 0.75 |
| 0.86 | 1 | 142: Fitfulness: irregularity of … | No | 3 | 18 | 71 | 1.00 |
| 0.86 | 5 | 576: Inelegance | Yes | 2 | 36 | 118 | 1.00 |
| 0.85 | 1 | 3: Substantiality | Yes | 4 | 28 | 95 | 0.00 |
| 0.85 | 2 | 234: Edge | Yes | 5 | 35 | 149 | 0.80 |
| 0.84 | 1 | 23: Prototype | Yes | 5 | 34 | 121 | 0.60 |
| 0.84 | 3 | 251: Conduit | Yes | 2 | 27 | 93 | 1.00 |

**Table 2:** Distribution of words and phrases within *Roget's Thesaurus* ordered by percentage of common strings

define it. *SG* stands for "semicolon group".

Both lexical resources are similar in absolute size, containing about 200,000 words and phrases. 51% of all words in *Roget's* are nouns, 21% adjectives, 24% verbs, 3% adverbs and less than 1% are interjections. 67% of *WordNet* are nouns, 18% adjectives, 12% verbs and 3% adverbs (Jarmasz and Szpakowicz, 2001b). There are no interjections in *WordNet*. We expected a large overlap between both resources, but the 45,000 common words and phrases only represent about 41% of the unique words and phrases in the *Thesaurus* and 36% of *WordNet*. If we allow for repetitions, the common strings make up 63% of *Roget*. The equivalent calculation has not been performed for *WordNet*. We expected the overlap to be concentrated in the first three *Roget* classes, but the results in Table 1 show that the common strings are distributed pretty evenly across the whole resource. A head-per-head analysis shows that 77% of head names as well as 60% of keywords can be found in *WordNet*. 88% of heads have at least 50% of their words in common with those of *WordNet* and 74% of heads have at least 50% of keywords in common. Table 2 shows the 10 heads with the highest percentage of common strings. *H in WN* indicates that the head name can be found in *WordNet.*

Identifying the areas where the *Thesaurus* and *WordNet* overlap is of interest to us since we want to import semantic relations from *WordNet*. It uses a set of about 15 semantic relations (Miller *et al.* 1990), all of which are present in *Roget* but none of which are labelled in an explicit manner. In the next section we discuss how we can import automatically these relations.

## 3 Automatic Labelling of Semantic Relations in *Roget's Thesaurus*

The *Thesaurus* has a vast set of semantic relations but they are not labelled explicitly. This is a major hindrance for NLP applications: "*Roget's* remains … an attractive lexical resource for those with access to it. Its wide, shallow hierarchy is densely populated with nearly 200,000 words and phrases. The relationships among the words are also much richer than *WordNet's* IS-A or HAS-PART links. The price paid for this richness is a somewhat unwieldy tool with ambiguous links" (McHale, 1998). We propose two ways to deal with this lack of information: a similarity measure can be calculated for every pair of semicolon groups and, when possible, labelling of links based on *WordNet*.

McHale (ibid.) has shown that edge counting is a very good measure for calculating semantic similarity in the *Thesaurus*. In an experiment with 28 noun pairs he found that edge counting using *Roget's International Thesaurus* gives almost as good results as human judges. *WordNet* gives poor results with this metric. This is most probably due to the fact that *WordNet's* taxonomy is of varying depth, whereas *Roget's* is fixed with a maximum of 7 edges. (Jarmasz and Szpakowicz, 2001b). The advantages of edge counting, beyond the fact that it seems to be an accurate metric, are that it is easy to compute and can be calculated for all semicolon groups. Although not yet implemented, it will definitely be available through our software.

We plan to label as many semantic relations as possible in *Roget's*, but as a first step, only relations from a keyword to the semicolon groups within the same paragraph will be identified. Once this has been completed, we will start to look for relations across paragraphs. Kwong (1998) has suggested that it is possible to map *Roget's* and *WordNet* senses by identifying paragraphs and *mini-nets* that have mutiple-word overlap. A mini-net is a group of words and phrases found by following one link in all possible semantic relations for a word. Let us illustrate with an example, comparing the paragraph with keyword *decrement* to a mini-net for the noun *decrement*. The *Thesaurus* paragraph is the following:

Class one: *Abstract relations*
Section three: *Substantiality*
Head 42: *Decrement: thing deducted*

**N.** *decrement*, deduction, depreciation, cut 37 *diminution*; allowance; remission; tare, drawback, clawback, rebate, 810 *discount*; refund, shortage, slippage, defect 307 *shortfall*, 636 *insufficiency*; loss, sacrifice, forfeit 963 *penalty*; leak, leakage, escape 298 *outflow*; shrinkage 204 *shortening*; spoilage, wastage, consumption 634 *waste*; subtrahend, rake-off, 786 *taking*; toll 809 *tax*.

The mini-net for the noun *decrement* can be built in the following way:

<u>Overview of noun decrement</u>
The noun decrement has 2 senses:
1. decrease, decrement -- (the amount by which something decreases)
2. decrease, decrement -- (a process of becoming smaller)

<u>Synonyms/Hypernyms of noun decrement</u>
**Sense 1 -** decrease, decrement: *{amount}*
**Sense 2 -** decrease, decrement: *{process}*

<u>Hyponyms of noun decrement</u>
**Sense 1 -** decrease, decrement: *{drop, fall}*, *{shrinkage}*
**Sense 2 -** decrease, decrement: *{wastage}*, *{decay, decline}*, *{slippage}*, *{decline, diminution}*, *{desensitization, de sensitisation}*, *{narrowing}*

<u>Coordinate Terms of noun decrement</u>
**Sense 1 -** decrease, decrement : *{amount}*, *{quantity}*, *{increase, increment}*, *{decrease, decrement}*, *{insufficiency, inadequacy, deficiency}*, *{number, figure}*
**Sense 2 -** decrease, decrement : *{process}*, *{natural process, natural action, action, activity}*, *{photography}*, *{chelation}*, *{human process}*, *{development, evolution}*, *{economic process}*, *{decrease, decrement}*, *{increase, increment, growth}*, *{processing}*, *{execution}*, *{degeneration}*, *{shaping, defining}*, *{dealignment}*, *{uptake}*

By matching semicolon groups and synsets where at least one word or phrase is in common, it is possible to rearrange the *Roget* paragraph in the following manner :

**N.** *decrement*
**Hyponym** : deduction, depreciation, cut 37 *diminution*; refund, shortage, slippage, defect 307 *shortfall*, 636 *insufficiency*; shrinkage 204 *shortening*; spoilage, wastage, consumption 634 *waste*.
**No label** : allowance; remission; tare, drawback, clawback, rebate, 810 *discount*; loss, sacrifice, forfeit 963 *penalty*; leak, leakage, escape 298 *outflow*; subtrahend, rake-off, 786 *taking*; toll 809 *tax*.

Even though it is not significant, many interesting observations can be drawn from this simple experiment. Although 35 of the 36 words and phrases of the paragraph, and the synset {*leak*, *leakage*, *escape*, *outflow*} can be found in *WordNet*, only 6 of these words were found in the *decrement* mini-net. This leads us to believe that, even when both resources use the same lexical material, their organization is much different. We see that it is possible to import semantic relations from *WordNet* and the mapping of senses can be seen as a way of reorganizing the senses within *Roget's* ontology. It may also be possible to increase *WordNet's* lexical material using words and phrases from *Roget's* via this mapping, but there is no automatic manner of evaluating the validity of the new synset.

### Conclusion

We have implemented an ELKB using *Roget's Thesaurus*. Although the software is not entirely ready for large-scale use, we propose realistic solutions to the problem of identifying semantic relations in the *Thesaurus*. We have shown that although *WordNet* and *Roget's* are conceptually similar, they differ when it comes to lexical material and organization. A mapping between *Roget's* paragraphs and *WordNet* mini-nets makes it possible to exchange information between the two resources.

**Acknowledgements**

This research would not have been possible without the help of Pearson Education, the owners of the 1987 Penguin's *Roget's Thesaurus of English Words and Phrases*; Steve Crowdy and Martin Toseland helped obtain this lexical resource and answered our many questions. Partial funding for this work comes from the Natural Sciences and Engineering Research Council of Canada.